\documentclass[letterpaper, 10 pt, conference]{ieeeconf}

\IEEEoverridecommandlockouts                              

\usepackage{url} 
\usepackage{amsmath} 
\usepackage{amssymb}  
\usepackage{mathptmx} 
\DeclareMathAlphabet{\mathcal}{OMS}{cmsy}{m}{n}

\usepackage[ruled,vlined,noend]{algorithm2e}
\usepackage{footmisc}
\usepackage{bm}
\usepackage{multirow}
\usepackage{xcolor}

\newcommand{\bs}{\boldsymbol}

\newcommand{\new}[1]{\textcolor{black}{#1}}

\usepackage{tikz}
\usetikzlibrary{tikzmark,calc, decorations.pathreplacing}
\newcommand\mybrace[4]%
   {\begin{tikzpicture}[remember picture,overlay]
    \draw[decorate,decoration={brace,raise=440pt}]
      ([yshift=2ex]{{pic cs:#3}|-{pic cs:#1}}) --
      node[xshift=444pt,anchor=west] {#4} 
      ([yshift=-0.5ex]{{pic cs:#3}|-{pic cs:#2}});
    \end{tikzpicture}%
   }
\usepackage{array}

\usepackage{cite}

\title{\LARGE \bf
Policy Decomposition: Approximate Optimal Control with Suboptimality Estimates}

\author{Ashwin Khadke and Hartmut Geyer
\thanks{This work was supported by the NSF (grant \#1563807).}
\thanks{Ashwin Khadke and Hartmut Geyer are with the Robotics Institute, 
        Carnegie Mellon University, 5000 Forbes Avenue, Pittsburgh, PA 15213, USA
        {\tt\small \{akhadke, hgeyer\}@andrew.cmu.edu}}%
}

\begin{document}
\begin{table*}[t!]
\large{
This work has been submitted to the IEEE for possible publication. Copyright may be transferred without notice, after which this version may no longer be accessible.}
\end{table*}
\maketitle

\begin{abstract}
\new{Numerically computing global policies to optimal control problems for complex dynamical systems is mostly intractable.} In consequence, a number of approximation methods have been developed. However, none of the current methods can quantify by how much the resulting control underperforms the elusive globally optimal solution. Here we propose policy decomposition, an approximation method with explicit suboptimality estimates. Our method decomposes the optimal control problem into lower-dimensional subproblems, whose optimal solutions are recombined to build a control policy for the entire system. Many such combinations exist, and we introduce the value error and its LQR and DDP estimates to predict the suboptimality of possible combinations and prioritize the ones that minimize it. Using a cart-pole, a 3-link balancing biped and N-link planar manipulators as example systems, we find that the estimates correctly identify the best combinations, yielding control policies in a fraction of the time it takes to compute the optimal control without a notable sacrifice in closed-loop performance. While more research will be needed  
to find ways of dealing with 
the combinatorics of policy decomposition, the results suggest this method could be an effective alternative for approximating optimal control in intractable systems. 
\end{abstract}

\section{INTRODUCTION}

\new{Owing to the curse of dimensionality, obtaining global control policies for complex nonlinear systems common in robotics requires approximation methods to remain computationally tractable \cite{Bertsekas:1995}.} Several such methods exist, relying on either local search or state-space reduction. Local search methods focus on the behavior of a system close to a reference motion and iteratively update both, the control and the reference motion, to achieve a desired behavior. Primary examples of this category include open-loop trajectory optimization \cite{Von:1992} as well as closed-loop methods such as DDP \cite{Mayne:1973} and iLQG \cite{Todorov:2005}, and their recent extensions \cite{Todorov:2009, Tassa:2014, Farshidian:2017}. The resulting locally optimal solutions have been combined to construct global control policies \cite{Tedrake:2010, Atkeson:2013, Zhong:2013}. But, it remains open how closely these global policies approximate the true optimal control of the full system.

Approximation methods that reduce the state space to simplify the optimal control problem face a similar issue. These methods express the global control of a dynamical system as functions of its lower-dimensional features \cite{Tsitsiklis:1996}. The features can be hand designed \cite{Stilman:2005}, parameterized with basis functions \cite{Engel:2005}, or derived by minimizing some projection error \cite{Bouvrie:2017}. While these features sufficiently capture the original system dynamics, they are agnostic to the objective of the optimal control problem. As a result, the gap in performance between the resulting global control and the optimal control of the system remains difficult to assess. Some recent extensions try to overcome this issue. For instance, \cite{Xue:2016} and \cite{Alla:2017} identify subspaces within the state space for which the optimal control of the corresponding lower-dimensional system deviates little in closed-loop behavior from the optimal control of the entire system. However, these extensions only work for linear systems \cite{Xue:2016} or linear approximations of nonlinear systems \cite{Alla:2017}. Reduction methods that can assess the suboptimality of their resulting controls when applied to the original, nonlinear system remain elusive.

We propose policy decomposition, an approximate method for solving optimal control problems with suboptimality estimates for the resulting controllers. Policy decomposition builds on two main ideas. First, it breaks the optimal control problem for a complex system into lower-dimensional subproblems, from which it builds a control policy for the full system in a cascaded fashion. Second, it introduces the error between the value functions of control policies obtained with and without decomposition to measure the closed-loop performance of possible decompositions \emph{a priori}. 
 As this error cannot be computed without knowing the true optimal control, we estimate it based on LQR and DDP approximations. Choosing one or the other estimate trades off computational speed (LQR) and prediction accuracy (DDP). We first overview the main ideas behind policy decomposition (Sec.~\ref{sec:Overview}) and then develop them more formally (Sec.~\ref{sec:Decomposition}--\ref{sec:DDP}). Using a cart pole, a 3-link balancing biped, and N-link planar manipulators, we show the proposed method can find control policies in a fraction of the time it takes to solve the optimal control while sacrificing little in closed-loop performance. Finally, we discuss strategies for dealing with the combinatorics of policy decomposition  (Sec.~\ref{sec:discussion}).
\begin{figure*}[t!]
\centering
\includegraphics[width=\textwidth]{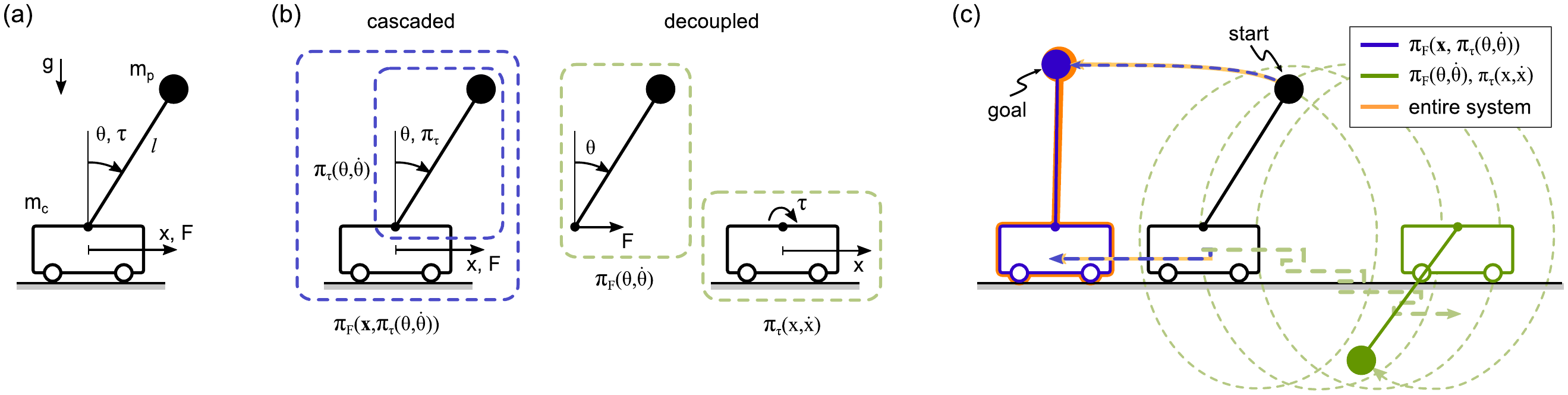}
\caption{Control of cart-pole system using policy decomposition. (\textbf{a}) Cart-pole model. See section~\ref{sec:Overview} for definitions and appendix B for details. (\textbf{b}) Cascaded and decoupled examples of policy decomposition.  (\textbf{c}) Resulting closed-loop behavior (blue and green traces) in comparison to optimal control of entire system (red) given model parameters $m_c = 5$kg, $m_p = 1$kg, and $l = 0.9$m, and bounds on the inputs, $|F| \leq 6$N and $|\tau| \leq 6$Nm.}
\label{fig:cartpole}
\end{figure*}
\section{Overview} \label{sec:Overview}
Consider designing a control policy to swing-up a pole on a cart while moving the cart to a goal position (Fig.~\ref{fig:cartpole}). The dynamics of this cart-pole system are given by
\begin{equation} \label{eq:CartPoleDynamics}
 \begin{split}
 \ddot{x} = &\frac{F - \frac{\tau}{l}\cos\theta + m_pl\dot{\theta}^2\sin\theta + \frac{m_p g}{2}\sin 2\theta}{m_c + m_p\sin\theta}\\
\ddot{\theta} = &\frac{ \frac{\tau}{l^2} (\frac{m_c}{m_p} + 1) - \frac{F}{l}\cos\theta - \frac{m_p \dot{\theta}^2}{2}\sin 2\theta 
- \frac{g}{l} (m_c + m_p)  \sin\theta } { m_c + m_p\sin\theta}
\end{split}
\end{equation}
where $x$ and $\dot{x}$ are the horizontal position and velocity of the cart, $\theta$ and $\dot{\theta}$ are the angle and angular velocity of the pole, and the cart force $F$ and pole torque $\tau$ are the two control inputs driving the system. In addition, the model parameters $m_p$ and $m_c$ are the masses of the pole and cart, respectively, $l$ is the pole length, and $g$ is the gravitational acceleration (Fig.~\ref{fig:cartpole}-a). Although with a six-dimensional state-action space the example is simple and its control optimization tractable, imagine it were not. To simplify the optimization problem, one could first optimize an inner control policy $\pi_{\tau}(\theta, \dot{\theta})$ for $\tau$ to swing up the pole assuming the cart is locked and then optimize an outer policy $\pi_{F}\left(x,\dot{x},\theta,\dot{\theta},\pi_{\tau}(\theta,\dot{\theta})\right)$ for $F$ to move the cart with the torque control of the pole set to $\pi_{\tau}$. An alternative to this cascade, is to treat the cart and pole as decoupled subsystems and design independent control policies $\pi_{\tau}(x,\dot{x})$ and $\pi_{F}(\theta, \dot{\theta})$ (Fig.~\ref{fig:cartpole}-b). Both policy decompositions, and the other 42 possible ones, reduce the dimensionality of the problem and make it computationally much more tractable (computational speed gains of one to three orders of magnitude). But the quality of the resulting control differs considerably among the decompositions. For instance, the cascaded policy optimization suggested first (blue trace, Fig.~\ref{fig:cartpole}-c) performs about as well as the true optimal control (red reference trace). On the other hand, the decoupled policy decomposition example performs much worse; in fact, it never reaches the goal state (green trace). 

Some of the performance outcomes seem intuitive. For instance, the pole dynamics are independent of the position and velocity of the cart (Eq.~\ref{eq:CartPoleDynamics}), which suggests the cascaded control optimization with an inner policy $\pi_{\tau}(\theta,\dot{\theta})$ disregarding the cart should perform well. 
For more complex dynamical systems, however, intuition quickly fades, and it gets difficult to predict the closed-loop performance of any control decomposition \cite{Arranz:2017} including policy decomposition.

To measure the quality of closed-loop behavior for a decomposition $\delta$, we introduce the value error, $\text{err}^\delta$, defined as the average difference between the value functions $V^\delta$ and $V^*$ of control policies obtained with and without decomposition,
\begin{equation}
    \text{err}^\delta = \frac{1}{|\mathcal{S}|}\: \int_\mathcal{S} V^\delta(\boldsymbol{x}) - V^*(\boldsymbol{x}) \:\: d\boldsymbol{x}
    \label{eq:errpolicy}
\end{equation}
where $\mathcal{S}$ is the state space and $\boldsymbol{x} \in \mathcal{S}$. As defined, $\text{err}^\delta$ directly quantifies the suboptimality of the resulting control induced by a decomposition. Because this measure cannot be computed without knowing the true value function $V^\ast$ of the original and intractable optimal control problem, we explore two methods of estimating the value error. In the first method, we linearize the system dynamics and estimate $\text{err}^\delta$ using the LQR solutions \cite{Kwakernaak:1972} for the complete linear system and its equivalent policy decompositions. These estimates $\text{err}^\delta_\text{lqr}$ are very fast to compute but loose accuracy away from the point of linearization. Alternatively, we compute DDP solutions \cite{Mayne:1973} for the original and decomposed systems from a few initial states to estimate $V_\delta$, $V^*$, and the value error. This second method improves the accuracy of the error estimate, $\text{err}^\delta_\text{ddp}$, but largely increases the computational costs.

\begin{figure}[t!]
    \centering
    \includegraphics[width=1\columnwidth]{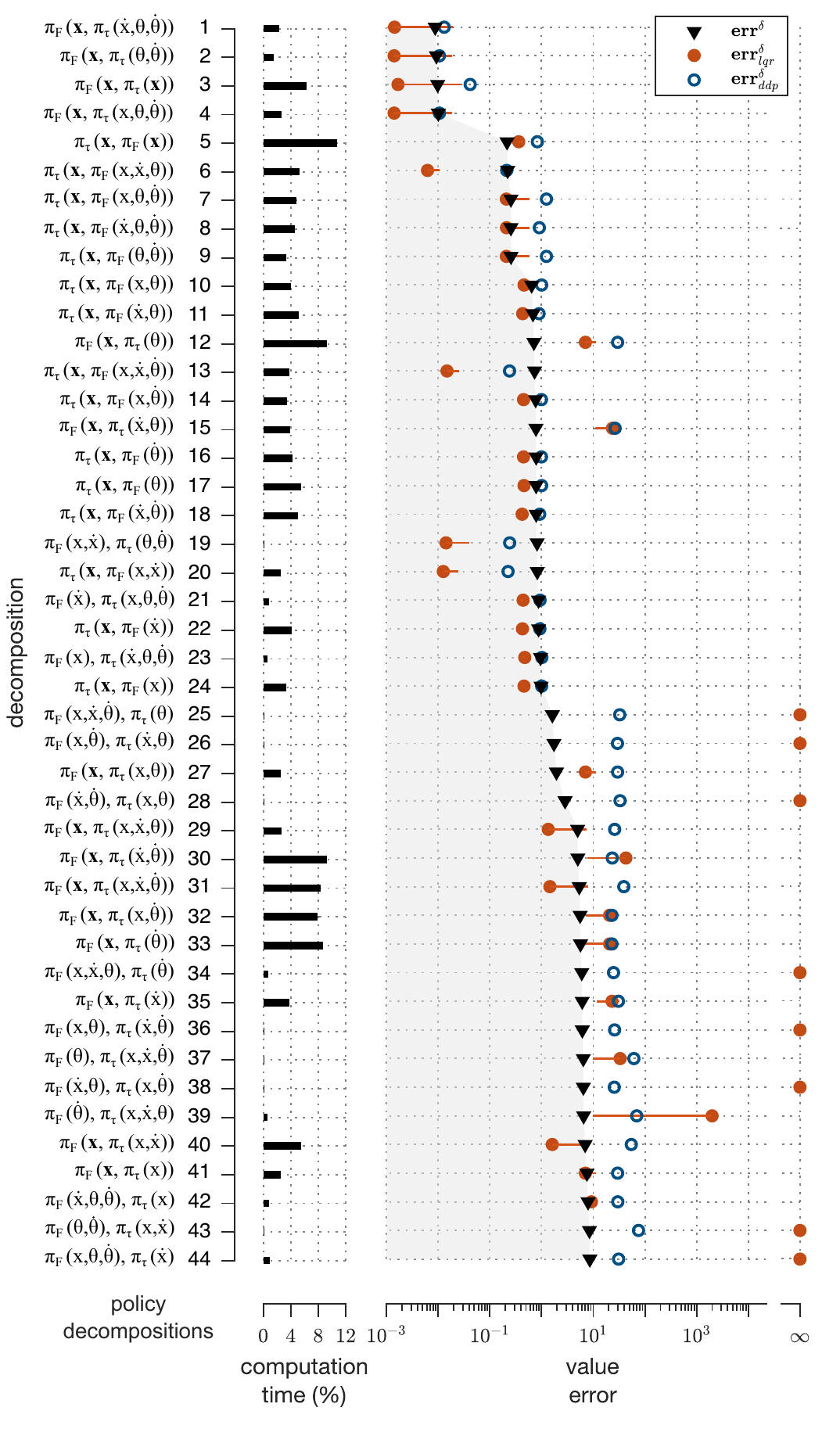}
    \caption{Cart-pole policy decompositions. The computation times (relative to optimal control) and value errors of all decompositions (triangles) are shown together with their LQR and DDP estimates (filled and open circles). LQR estimates include an error bar (solid lines) assessing the effect of ignored input bounds and are set to infinity for uncontrollable systems after linearization.}
    \label{fig:ValueErrorResults}
\end{figure}

The actual and predicted closed-loop performances for all policy decompositions of the cart pole system are summarized in Fig~\ref{fig:ValueErrorResults}. Both estimates (filled and open circles) correctly predict the four best performing policy decompositions, which have virtually the same value error of about 0.01 (triangles). The four decompositions include the cascaded policy described before (compare Fig.~\ref{fig:cartpole}, ranked $2^{\text{nd}}$ in Fig.~\ref{fig:ValueErrorResults}). It has the lowest complexity (lowest-dimensional state-action space for inner policy, same state-action space for outer policy) and is fastest to compute among the four, yielding a solution to the cart-pole control problem 66 times faster than computing the true optimal control. In summary, an algorithm using either estimate could have identified this decomposition \emph{a priori} and then computed the corresponding control policy in a fraction of the time it takes to compute the true optimal control policy without notably sacrificing closed-loop performance (blue vs. red trace, Fig.~\ref{fig:cartpole}-c).

\section{Policy Decompositions} \label{sec:Decomposition}

To formally develop the idea of policy decomposition, we consider the general dynamical system 
\begin{equation} \label{eq:dynamics}
\dot{\bs{x}} = \bs{f}(\bs{x}, \boldsymbol{u})
\end{equation} 
with state $\bs{x}$ and input $\bs{u}$. The optimal control for this system is defined as the control policy $\pi^*_{\bs{u}}(\bs{x})$ that minimizes

\begin{equation} \label{eq:cost}
J_{} = \int_0^\infty e^{-\lambda t} c(\bs{x}(t), \bs{u}(t)) \: dt.
\end{equation}
This objective function describes the discounted sum of some costs $c(\bs{x},\bs{u})$ accrued over time with the discount factor $\lambda$ characterizing the trade-off between immediate and future costs. We assume a quadratic structure of $c(\bs{x},\bs{u})$, 
\begin{equation} \label{eq:obj}
c(\bs{x}, \bs{u}) = (\bs{x} - \bs{x}^d)^T \bs{Q} (\bs{x} - \bs{x}^d) + (\bs{u}-\bs{u}^d)^T \bs{R} (\bs{u}-\bs{u}^d)
\end{equation}
where $\bs{x}^d$ and $\bs{u}^d$ define the goal state and input. To approximate the optimal control, policy decomposition reduces the search for one high-dimensional policy to a search for a collection of sub-policies that are lower-dimensional and much faster to compute. In general, the number of possible decompositions grows in a combinatorial manner with the dimensions of the state and input vectors. At a fundamental level, however, these combinations are comprised of only two building blocks: decoupled and cascaded sub-policies. 

A \emph{decoupled} sub-policy $\pi_{\bs{u}_i}(\bs{x}_i)$ is the optimal control for the subsystem
\begin{equation} \label{eq:decdyn}
\dot{\bs{x}}_i = \bs{f}_i (\bs{x}_i, \bs{u}_i \mid\: \bar{\bs{x}}_i=\bar{\bs{x}}_i^d, \bar{\bs{u}}_i=\bs{0})
\end{equation}
where $\bs{x}_i$ and $\bs{u}_i$ are subsets of $\bs{x}$ and $\bs{u}$, $\bs{f}_i$ only contains the dynamics associated with $\bs{x}_i$ (Eq.~\ref{eq:dynamics}), and the complement state and input vectors, $\bar{\bs{x}}_i = \bs{x} \setminus \bs{x}_i$ and  $\bar{\bs{u}}_i = \bs{u}\setminus\bs{u}_i$, are assumed to be constant parameters. Specifically, $\bar{\bs{u}}_i=\bs{0}$ decouples any influence from complement inputs on the subsystem dynamics. The corresponding cost function reduces to
\begin{equation} \label{eq:deccost}
c_i^{\text{dec}}(\bs{x}_i,\bs{u}_i) = (\bs{x}_i - \bs{x}_i^d)^T \bs{Q}_i (\bs{x}_i - \bs{x}_i^d) + (\bs{u}_i-\bs{u}_i^d)^T \bs{R}_i (\bs{u}_i-\bs{u}_i^d)
\end{equation}
where $\bs{Q}_i$ and $\bs{R}_i$ are appropriate sub-matrices of $\bs{Q}$ and $\bs{R}$.

A \emph{cascaded} sub-policy $\pi_{\bs{u}_i}(\bs{x}_i,  \pi_{\bar{\bs{u}}_i}(\bs{x}_i))$, on the other hand, is the optimal control for the subsystem
\begin{equation} \label{eq:cascdyn}
\dot{\bs{x}}_i = \bs{f}_i (\bs{x}_i, \bs{u}_i \mid\: \bar{\bs{x}}_i=\bar{\bs{x}}_i^d, \bar{\bs{u}}_i=\pi_{\bar{\bs{u}}_i}(\bs{x}_i)).
\end{equation}
This subsystem differs from the decoupled one in the complement input, which becomes an inner sub-policy with at least one non-zero element,
\begin{equation} \label{eq:cascadeinput}
    \pi_{\bar{\bs{u}}_i}(\bs{x}_i) = [0,\: \ldots,\: 0,\: \pi_{\bs{u}_j}(\bs{x}_j), \:0 ,\: \ldots,\: 0]
\end{equation}
where $\bs{u}_j\subseteq\bar{\bs{u}}_i$ and $\bs{x}_j\subseteq\bs{x}_i$. Note that (i) $\pi_{\bar{\bs{u}}_i}(\bs{x}_i)$ can contain multiple sub-policies, (ii) these sub-policies can be either cascaded (not shown in Eq.~\ref{eq:cascadeinput}) or decoupled ones, and (iii) they have to be known before $\pi_{\bs{u}_i}(\bs{x}_i,  \pi_{\bar{\bs{u}}_i}(\bs{x}_i))$ can be computed. In addition, the cost function changes to 
\begin{equation} \label{eq:CasCost}
\begin{split}
   c_i^{\text{cas}}(\bs{x}_i,\bs{u}_i)  = & (\bs{x}_i - \bs{x}_i^d)^T \bs{Q}_i (\bs{x}_i - \bs{x}_i^d) + (\bs{u}_i-\bs{u}^d_i)^T \bs{R}_i (\bs{u}_i-\bs{u}^d_i) \\
   &   + (\bs{u}_j-\bs{u}_j^d)^T \bs{R}_j (\bs{u}_j-\bs{u}_j^d).
\end{split}
\end{equation}

Purely decoupled policy decompositions are the fastest to compute whereas cascaded ones tend to offer better closed-loop performance. 
For instance, a system with $n$-dimensional state and $m=n$ inputs requires jointly computing $n$-dimensional policies for all the inputs. By contrast, a purely decoupled decomposition,
\begin{equation} \label{eq:DecPolicy}
\pi_{\bs{u}}^{\text{dec}}(\bs{x}) = \left(\pi_{u_1}(x_1), \ldots, \pi_{u_n}(x_n)\right)
\end{equation}
resolves to $n$ 1-dimensional policies to be computed independently. The associated reduction in computation time is dramatic but may be bought at the cost of a poor closed-loop performance, precisely because decoupling ignores cross-influences. Cascaded decompositions partly include these influences and tend to offer better performance. However, they generate optimal control problems that quickly become more complex and computationally costly. 
A purely cascaded decomposition for such a system,
\begin{equation} \label{eq:PureCascadeDecomp}
\begin{split}
\pi^{\text{cas}}_{\bs{u}}(\bs{x}) =& \big(\pi_{u_1}(x_1), \pi_{u_2}((x_1, x_2), \pi_{u_1}(x_1)), \:\ldots, \\
& \:\:\: \pi_{u_n}(\bs{x}, \pi_{u_{n-1}}((x_1, \: \ldots, \:x_{n-1}), \pi_{u_{n-2}}(\:\ldots, \: \ldots)))\big) 
\end{split}
\end{equation}
results in computing $n$ policies that grow in dimensionality from 1 to $n$.

A general policy decomposition can combine decoupled and cascaded policies in many ways, posing a combinatorial challenge that requires one to curtail the number of decompositions being tested. For instance, counting just purely decoupled and cascaded decompositions for a system with $n$ states and $m$ inputs leads to a total of
 \begin{equation}
 \label{eq:combinations}
     N(n,m) = \sum_{r=2}^{m} \:\Delta(m, r) \left[\frac{\Delta(n, r)}{r!} + \big(r^n - (r-1)^n\big) \:\right]   
 \end{equation}
 possible combinations (see appendix A for details), where
 \begin{align} \label{eq:CombDelta}
 \Delta(a,b) = \sum_{k=0}^{b-1} (-1)^k \binom{b}{k} (b-k)^a
 \end{align}

The cart-pole system 
with two inputs and four states (Fig.~\ref{fig:cartpole}-a), has only 44 possible policy decompositions and value functions of all decomposed policies and the cart-pole's optimal control can be readily computed (implementation details in appendix B). 
They are ranked by the value error in Fig~\ref{fig:ValueErrorResults} (black triangles). Several of these decompositions greatly reduce computation time without giving up closed-loop performance. The four best ones have an error sufficiently small ($\text{err}^\delta \approx 0.01$) to virtually match the closed-loop performance of the optimal control (illustrated for the second ranked decomposition in Fig~\ref{fig:cartpole}-c). Yet the time to compute these control policies reduces by a factor of 16 (decomposition \#3) to 66 (\#2).

\section{LQR Suboptimality Estimate} \label{sec:LQR}
While the value error $\text{err}^\delta$ provides a measure for the suboptimality of policy decompositions $\delta$, it cannot be computed directly in intractable systems, as it requires to know the system's optimal control. We thus explore two methods of estimating the error.

The first method relies on the corresponding linear system, 
\begin{equation} \label{eq:LQRdyn}
    \dot{\bs{x}} = \bs{A}(\bs{x} - \bs{x}^d) + \bs{B}(\bs{u} - \bs{u}^d)
\end{equation}
obtained by linearizing the dynamics (\ref{eq:dynamics}) about the goal state and input,
\begin{equation}\label{eq:LQRapproxAB}
\bs{A} = \left.\frac{\partial\bs{f}(\bs{x}, \bs{u})}{\partial\bs{x}}\right|_{(\bs{x}^d, \bs{u}^d)}, \:\:
\bs{B} = \left.\frac{\partial \bs{f}(\bs{x}, \bs{u})}{\partial \bs{u}}\right|_{(\bs{x}^d, \bs{u}^d)}.    
\end{equation} 
Because the costs are quadratic (Eq.~\ref{eq:obj}), the optimal control $\pi^*_{\bs{u}}(\bs{x})$ of this system is an LQR, whose value function $V_{\text{lqr}}^*(\bs{x})$ can be readily computed by solving the algebraic Riccati equation \cite{Palanisamy:2015}. The value error estimate of a decomposition $\delta$ then becomes
\begin{equation} \label{eq:valueerrorLQR}
    \text{err}^\delta_\text{lqr} = \frac{1}{|\mathcal{S}|}\: \int_\mathcal{S} V^\delta_\text{lqr}(\boldsymbol{x}) - V^*_\text{lqr}(\boldsymbol{x}) \:\: d\boldsymbol{x}
\end{equation}
where $V_{\text{lqr}}^{\delta}(\bs{x})$ is the value function for the equivalent decomposition of the linear system. 

Performing the equivalent policy decomposition amounts to computing LQR gain matrices for the equivalent subsystems. A decoupled sub-policy $\pi_{\bs{u}_i}(\bs{x}_i)$ is replaced by the optimal control for the subsystem
\begin{equation} \label{eq:LQRdecdyn}
    \dot{\bs{x}}_i = \bs{A}_i (\bs{x}_i - \bs{x}_i^d) + \bs{B}_i (\bs{u}_i - \bs{u}_i^d)
\end{equation}
with
\begin{equation}
\bs{A}_i = \left.\frac{\partial \bs{f}_i}{\partial \bs{x}_i}\right|_{(\bs{x}^d,\bs{u}_i^d,\bar{\bs{u}}_i=\bs{0})}, \:\:
\bs{B}_i = \left.\frac{\partial \bs{f}_i}{\partial \bs{u}_i}\right|_{(\bs{x}^d,\bs{u}_i^d,\bar{\bs{u}}_i=\bs{0})}.
\end{equation}
Note that the assumptions about the complement state and input of a decoupled subsystem (Eq.~\ref{eq:decdyn}) are embedded by linearizing about the point $(\bs{x}, \bs{u}_i, \bar{\bs{u}}_i)=(\bs{x}^d, \bs{u}_i^d, \bs{0})$. The optimal control resolves to $\pi_{\bs{u}_i}(\bs{x}_i) = \bs{u}_i^d-\bs{K}_i (\bs{x}_i-\bs{x}_i^d)$, where $\bs{K}_i$ is the corresponding LQR gain matrix. Similarly, a cascaded sub-policy $\pi_{\bs{u}_i}(\bs{x}_i,  \pi_{\bar{\bs{u}}_i}(\bs{x}_i))$ becomes the LQR control for the subsystem 
\begin{equation} \label{eq:LQRcasdyn}
    \dot{\bs{x}}_i = (\bs{A}_i + \bs{\Pi}_i) (\bs{x}_i - \bs{x}_i^d) + \bs{B}_i (\bs{u}_i - \bs{u}_i^d)  
\end{equation}
with
\begin{equation}
\bs{A}_i = \left.\frac{\partial \bs{f}_i}{\partial \bs{x}_i}\right|_{\left(\bs{x}^d,\bs{u}_i^d,\bar{\bs{u}}_i=\pi_{\bar{\bs{u}}_i}(\bs{x}_i^d)\right)},
\bs{B}_i = \left.\frac{\partial \bs{f}_i}{\partial \bs{u}_i}\right|_{\left(\bs{x}^d,\bs{u}_i^d,\bar{\bs{u}}_i=\pi_{\bar{\bs{u}}_i}(\bs{x}_i^d)\right)}.
\end{equation}
This subsystem features the term $\bs{\Pi}_i (\bs{x}_i - \bs{x}_i^d)$, that embeds the inner sub-policy, $\bar{\bs{u}}_i = \pi_{\bar{\bs{u}}_i}(\bs{x}_i)$ (Eq.~\ref{eq:cascdyn}). Similar to (\ref{eq:cascadeinput}), $\bs{\Pi}_i$ contains at least one non-zero element, for instance,
\begin{equation} \label{eq:LQRPi}
    \bs{\Pi}_i  = \left[
        \begin{array}{cccc}
            0 & 0 & \cdots & 0 \\
            0 & -\overbrace{\bs{B}_j\bs{K}_j\rule{0cm}{0.3cm}}^{\text{dim}(\bs{x}_j)} &  & \vdots\\
            \vdots &  & \ddots & \vdots\\
            0 & \cdots & \cdots & 0
        \end{array}
        \right]     
\end{equation}
formed by the LQR gain $\bs{K}_j$ and the input matrix $\bs{B}_j$ of the inner sub-policy, $\pi_{\bs{u}_j}(\bs{x}_j) = \bs{u}_j^d-\bs{K}_j (\bs{x}_j-\bs{x}_j^d)$, with $\bs{x}_j\subseteq\bs{x}_i$ and $\bs{u}_j\subseteq \bar{\bs{u}}_i$.
In effect, the equivalent policy is a linear controller, 
\begin{equation}
\pi_{\bs{u}}^{\delta}(\bs{x}) = \bs{u}^d - \bs{K}^{\delta}(\bs{x} - \bs{x}^d)    
\end{equation}
whose gain $\bs{K}^\delta$ is a block matrix composed of all the subsystem LQR gains $\bs{K}_i$. More specifically, for the purely decoupled and cascaded decompositions with $r$ subsystems, the gain $\bs{K}^\delta$ takes on the general form
\begin{equation}
    \bs{K}^\text{dec} = \left[
        \arraycolsep=2.2pt\def\arrayhstretch{2.4} 
        \begin{array}{cccc}
            \overbrace{\bs{K}_1}^{\text{dim}(\bs{x}_1)} & 0 & \cdots & 0 \\
            0 & \overbrace{\bs{K}_2}^{\text{dim}(\bs{x}_2)} & \cdots & 0\\
            \vdots & \vdots & \ddots & \vdots\\
            0 & \cdots & 0 & \overbrace{\bs{K}_r}^{\text{dim}(\bs{x}_r)}
        \end{array}
        \right]
\end{equation}
and
\begin{equation} \label{eq:Kcasc}
    \bs{K}^\text{cas} = \left[
    \arraycolsep=2.2pt\def\arrayhstretch{2.0}
    \begin{array}{cccc}
        \multicolumn{1}{c}{\overbrace{\bs{K}_1}^{\text{dim}(\bs{x}_1)}} & 0 & \cdots & 0 \\
        \multicolumn{2}{c}{\overbrace{\rule{1.8cm}{0pt}}^{\text{dim}(\bs{x}_2)}} & & \\
         \bs{K}_2& & \cdots & 0\\
        \vdots & & \ddots & \vdots\\
        \multicolumn{4}{c}{\overbrace{\rule{3.6cm}{0pt}}^{\text{dim}(\boldsymbol{x})}} \\
        \bs{K}_r& & & \\
    \end{array}\right]
\end{equation}
respectively. With the gain $\bs{K}^\delta$ defined, the value function of an equivalent linear system decomposition $\delta$ resolves to
\begin{equation} \label{eq:LQRVdelta}
V_{\text{lqr}}^{\delta}(\bs{x}) = (\bs{x} - \bs{x}^d)^T\bs{P}^{\delta}(\bs{x} - \bs{x}^d) 
\end{equation}
where $\bs{P}^\delta$ is the solution of the Lyapunov equation, 
\begin{equation}
  \left(\bs{A} - \bs{B}\bs{K}^{\delta} - \frac{\lambda\bs{I}}{2} \right)^T \bs{P}^{\delta} \left(\bs{A} - \bs{B}\bs{K}^{\delta} - \frac{\lambda\bs{I}}{2} \right) + \bs{Q} + {\bs{K}^{\delta}}^T\bs{R}\bs{K}^{\delta} = 0.
\end{equation}

    The LQR suboptimality estimate can be computed within minimal time, but it 
    has some drawbacks. 
    \new{First, it only accounts for linearized system dynamics at the goal state. Second, the estimate is agnostic to bounds on the control inputs present in the original optimal control problem. Third, controllers obtained by decomposing the equivalent linear system may be closed-loop unstable ($\bs{P}^\delta$ in Eq.~(\ref{eq:LQRVdelta}) has negative eigenvalues) resulting in $\text{err}^\delta_\text{lqr}=\infty$.} 
    For the cart pole, the LQR suboptimality estimate (filled circles) broadly predicts the observed closed-loop performance of almost half of the policy decompositions (triangles) as can be seen in Fig.~\ref{fig:ValueErrorResults}. However, clear deviations occur. The severity of the resulting deviations is indicated by the error bars in Fig~\ref{fig:ValueErrorResults} (red solid lines), which show the difference between the LQR estimate and the value error obtained after applying the LQR control policies to the original cart-pole dynamics with the input bounds enforced. 

\section{DDP Suboptimality Estimate} \label{sec:DDP}
We explore a second method based on DDP \cite{Tassa:2014} to estimate the value error (Eq. (\ref{eq:errpolicy})). In contrast to LQR, DDP can enforce input bounds and account for system dynamics away from the goal state. However, these benefits have to be bought with a costly increase in computation time. 

DDP optimizes the closed-loop performance of a system about an initial reference trajectory $\bs{X}^0(t)$ generated from an input guess $\bs{U}^0(t)$. For every point $t$ along this time trajectory, DDP uses approximate system dynamics 
to iteratively update the input, $\bs{U}^+(t) = \bs{U}^-(t)-\bs{K}(t) \left(\bs{x}-\bs{X}^-(t)\right)$, and resulting trajectory, $\bs{X}^+(t)$, such that the cost (\ref{eq:cost}) is minimized. In effect, DDP produces a locally optimal solution, $\bs{X}(t)$ and $\bs{U}(t)$, whose value function,
\begin{equation}
V_{\text{ddp}}(\bs{x}) = \int_0^{t_{\max}} e^{-\lambda t} c\left(\bs{X}\left(t\right), \bs{U}(t)\right) dt
\end{equation}
approximates $V^*(\bs{x})$ for the system under consideration at the point $\bs{x}=\bs{X}(0)$ in the state space.

\begin{figure}[t!]
    \centering
    \includegraphics[width=\columnwidth]{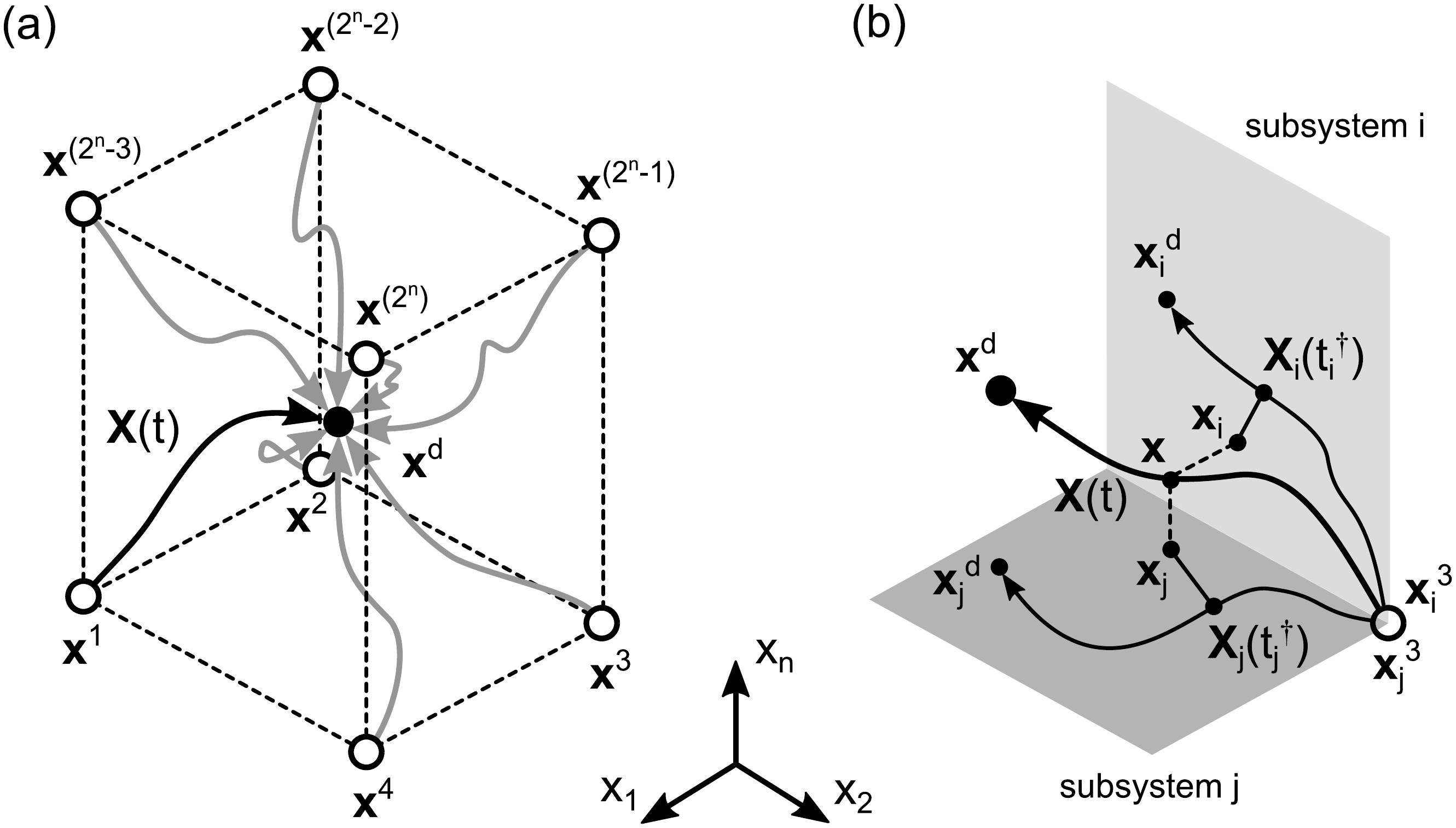}
    \caption{DDP approximation of value function. (\textbf{a}) Local DDP solutions $\bs{X}(t)$ for $k=2^n$ initial points $\bs{x}^s$ located at edges of hyper-cube that defines boundary of explored state space section. (\textbf{b}) Nearest neighbors $\bs{X}_i(t_i^\dagger)$ on subsystem solutions $\bs{X}_i(t)$ for current state $\bs{x}$ along solution $\bs{X}(t)$.}
    \label{fig:ddpmethod}
\end{figure}

We use this approximation to estimate the value error. Specifically, we introduce the suboptimality estimate 
\begin{equation} \label{eq:DDPvalueError}
    \text{err}_{\text{ddp}}^{\delta} = \frac{1}{k} \:\:\sum_{s=1}^k\:\big( V_{\text{ddp}}^{\delta}(\boldsymbol{x}^s) - V_{\text{ddp}}^*(\boldsymbol{x}^s)\big)
\end{equation}
which averages the value errors obtained from local DDP solutions to the original and decomposed optimal control problems for $k$ initial points centered on the goal state $\bs{x}^d$ (Fig.~\ref{fig:ddpmethod}-a). While $V_{\text{ddp}}^*(\boldsymbol{x}^s)$ can be computed right away, obtaining $V_{\text{ddp}}^{\delta}(\boldsymbol{x}^s)$ requires more interpretation. 

A policy decomposition $\delta$ with $r$ subsystems creates $r$ optimal control problems, whose individual DDP solutions need to somehow be combined for computing the approximate value function $V_\text{ddp}^\delta(\bs{x}^s)$. We achieve this with the following procedure. 
\new{First, starting from the initial sub-states $\{\bs{x}_i^s | s\in{1,\cdots,k}\}$ we use DDP to find for each subsystem $i$ locally optimal solutions characterized by $\bs{X}^{s}_i(t)$, $\tilde{\bs{X}}^{s}_i(t)$, $\bs{U}^{s}_i(t)$ and $\bs{K}^{s}_i(t)$. $\bs{X}^{s}_i(t)$ are the final DDP trajectories for the subsystem $i$ originating from state $\bs{x}^{s}_i$. $\tilde{\bs{X}}^{s}_i(t)$, $\bs{U}^{s}_i(t)$ and $\bs{K}^{s}_i(t)$ are the control reference trajectory, control inputs and local linear gains respectively that result in the subsystem following trajectory $\bs{X}^{s}_i(t)$.} Next, we define the subsystem control policy as the nearest neighbor policy \cite{Atkeson:2013},
\begin{equation} \label{eq:DDPSubPol}
    \pi_{\bs{u}_i}(\bs{x}_i) = \bs{U}^{\new{s^{\dagger}}}_i(\new{t^\dagger}) - \bs{K}^{\new{s^{\dagger}}}_i(\new{t^\dagger}) \left(\bs{x}_i-\tilde{\bs{X}}^{\new{s^{\dagger}}}_i(\new{t^\dagger})\right)
\end{equation}
where $\new{s^{\dagger}}$ and $\new{t^\dagger}$ respectively mark the trajectory ID and time at which $\bs{X}^{s}_i(t)$ is closest to the subsystem state (Fig.~\ref{fig:ddpmethod}-b), 
\begin{equation}
\new{s^{\dagger}, t^\dagger} = \text{arg}\min_{\new{s, t}} \left\Vert\bs{X}^{s}_i(t) - \bs{x}_i\right\Vert_2.
\end{equation}
Lastly, we run the policy $\pi^\delta_{\bs{u}}(\bs{x}) = \left(\pi_{\bs{u}_1}(\bs{x}_1),\ldots,\pi_{\bs{u}_r}(\bs{x}_r)\right)$  on the complete system (Eq. (\ref{eq:dynamics})) initialized at $\bs{x}^s$ and compute $V_{\text{ddp}}^\delta(\bs{x}^s)$ from the resulting trajectory $\bs{X}(t)$,
\begin{equation}
V_\text{ddp}^\delta(\bs{x}^s) = \int_0^{t_{\max}} e^{-\lambda t} c\left(\bs{X}\left(t\right), \pi_{\bs{u}}^\delta\left(\bs{X}(t)\right)\right) dt.
\end{equation}
Note that $\bs{X}(t)$ will differ from the collected trajectories of the individual DDP solutions, $\left(\bs{X}_1(t), \ldots, \bs{X}_r(t)\right)$, as the latter ignore at least some of the input couplings that influence the behavior of the complete system. 

DDP uses quadratic approximations of the system dynamics, but to curb computational costs, we consider only linear ones. The difference between a decoupled and a cascaded subsystem enters in this procedure through the approximate dynamics that DDP uses. Echoing the analysis presented in section~\ref{sec:LQR} (Eqs.~\ref{eq:LQRdecdyn}--\ref{eq:LQRPi}), the linearized subsystem dynamics at time $t$ along the trajectory $\bs{X}_i(t)$ resolve to
\begin{equation}
    \dot{\bs{x}}_i = \bs{A}_i^t \big(\bs{x}_i - \bs{X}_i(t)\big) + \bs{B}_i^t \big(\bs{u}_i - \bs{U}_i(t)\big)
\end{equation}
for a decoupled subsystem and 
\begin{equation}  \label{eq:DDPcasdyn}
    \dot{\bs{x}}_i = \bs{A}_i^t \big(\bs{x}_i - \bs{X}_i(t)\big) \new{+ \bs{\Pi}_i^t\big(\bs{x}_i - \tilde{\bs{X}}^t_i\big)} + \bs{B}_i^t \big(\bs{u}_i - \bs{U}_i(t)\big)
\end{equation}
for a cascaded one, with 
\begin{equation}\label{eq:DDPapproxAB}
\bs{A}_i^t = \left.\frac{\partial\bs{f}_i(\bs{x}_i, \bs{u}_i)}{\partial\bs{x}_i}\right|_{(\bs{X}_i(t), \bs{U}_i(t))}, \:\:
\bs{B}_i^t = \left.\frac{\partial \bs{f}_i(\bs{x}_i, \bs{u}_i)}{\partial \bs{u}_i}\right|_{(\bs{X}_i(t), \bs{U}_i(t))}   
\end{equation} 
As in Eq. (\ref{eq:LQRcasdyn}), the additional term $\new{\bs{\Pi}_i^t (\bs{x}_i - \tilde{\bs{X}}_i^t)}$ in the linearized dynamics of the cascaded subsystem (Eq. (\ref{eq:DDPcasdyn})) embeds at least one inner sub-policy $\pi_{\bs{u}_j}(\bs{x}_j)$, where LQR gain $\bs{K}_j$ \new{and desired state $\bs{x}^d_i$ (Eq. (\ref{eq:LQRPi})) are replaced by DDP gain $\bs{K}^{s_j^{\dagger}}_j(t_j^\dagger)$ and concatenation of reference states $\tilde{\bs{X}}^t_i = [\cdots, \tilde{\bs{X}}^{s_j^{\dagger}}_j(t_j^{\dagger}), \cdots]$ (Eq.~\ref{eq:DDPSubPol}) respectively. Note that $s_j^{\dagger}$ and $t_j^{\dagger}$ identify the nearest neighbour to $\bs{x}_i$ in trajectories of subsystem $j$. To generate the initial input sequence $\bs{U}^0(t)$ when computing a DDP policy 
for a decoupled subsystem $j$, we use the LQR controller gain $\bs{K}_j$ (described in section \ref{sec:LQR}) to roll-out trajectories and generate $\bs{U}^0(t)$. In case of a purely cascaded decomposition, we use $\bs{K}_j$ in conjunction with DDP policies (Eq.~\ref{eq:DDPSubPol}) of subsystems earlier in the cascade to compute initial trajectories.}

The DDP suboptimality estimate generally improves the value error prediction in the cart-pole example (Fig.~\ref{fig:ValueErrorResults}). Unlike the LQR estimate, the DDP estimate (open circles) 
does not suffer from the deviations due to uncontrollability and input bounds that affect the LQR estimate. But, computational costs for this improvement are high. For the cart pole system, computing the DDP estimate requires 60\% of the time it takes to actually compute the decomposed policy.

\section{Discussion} \label{sec:discussion}

We introduced policy decomposition, an approximate method for solving optimal control problems that reduces search for one high-dimensional control policy to a search for a collection of lower-dimensional sub-policies that are faster to compute yet preserve closed-loop performance when combined. We 
showed benefits of this idea with the cart-pole system 
 (Figs.~\ref{fig:cartpole} and \ref{fig:ValueErrorResults}).
Next, we introduced the value error (Eq. (\ref{eq:errpolicy})), a measure of a decomposition's suboptimality, and derived two estimates of it using LQR or DDP. The first estimate computes in minimal time while the second one improves the error prediction at the cost of added computation time. The estimates enable us to assess a decomposition's closed-loop performance without computing the policy.


A measure that predicts the closed-loop performance of control decompositions is a useful tool. Several measures have been proposed to help select simplified control configurations in complex systems. Measures using transfer functions \cite{Arranz:2017} and Gramians \cite{Conley:2000} build on open-loop dynamics, which may not correlate well with closed-loop behavior. 
 The $\nu$-gap measure \cite{Zhou:1998} overcomes this limitation 
 but ignores the objective of the underlying optimal control problem. Measures that account for the objective have been proposed for linear systems, including sum of output covariances of the resulting LQG control \cite{Halvarsson:2009} and value function bounds for LQR controllers obtained through nested$-\epsilon$ decompositions \cite{Siljak:2011}. In contrast, the value error (Eq. (\ref{eq:errpolicy})) is a general measure for any decomposition's suboptimality in nonlinear systems. The LQR (Eq. (\ref{eq:valueerrorLQR})) and DDP (Eq. (\ref{eq:DDPvalueError})) estimates we derived for policy decomposition may be adaptable to other control decompositions.


\begin{figure}[t]
  \centering
  \includegraphics[width=\columnwidth]{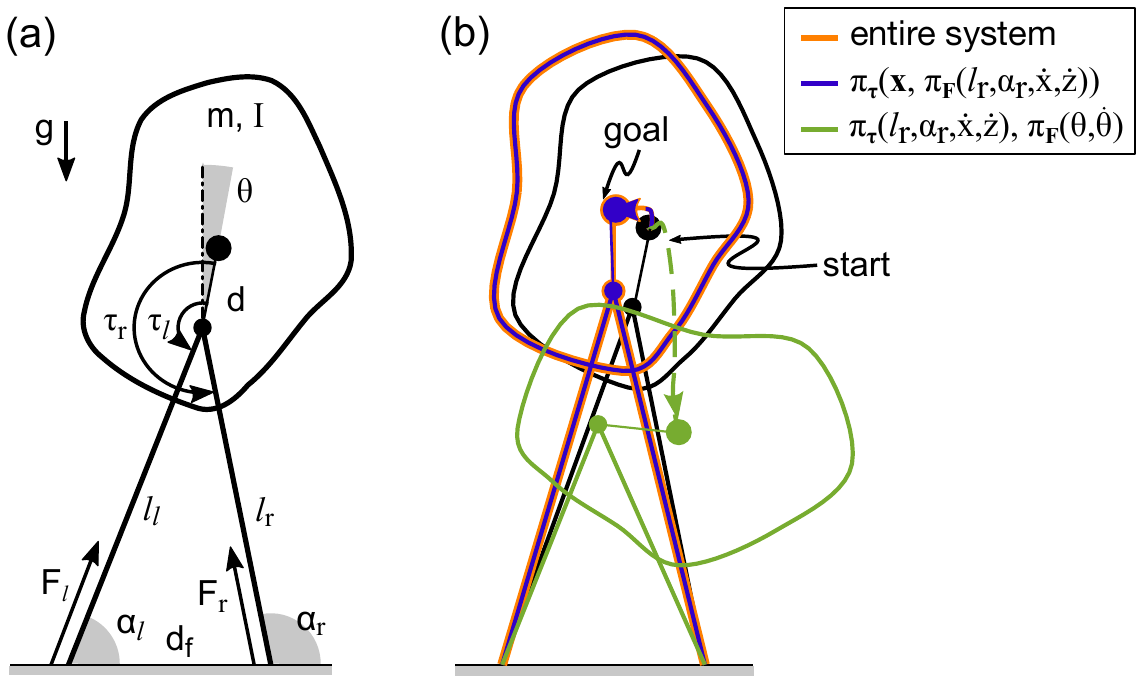}
  \caption{Balancing control for 3-link biped model. (\textbf{a}) Biped model. See section~\ref{sec:discussion} and appendix C for  definitions. (\textbf{b}) Behavior of optimal control (red traces) and best and worst policy decomposition (blue and green) (compare Tab.~\ref{tab:bipedresddp}).}
  \label{fig:biped}
\end{figure}

\begin{figure}[t]
  \centering
  \includegraphics[width=\columnwidth]{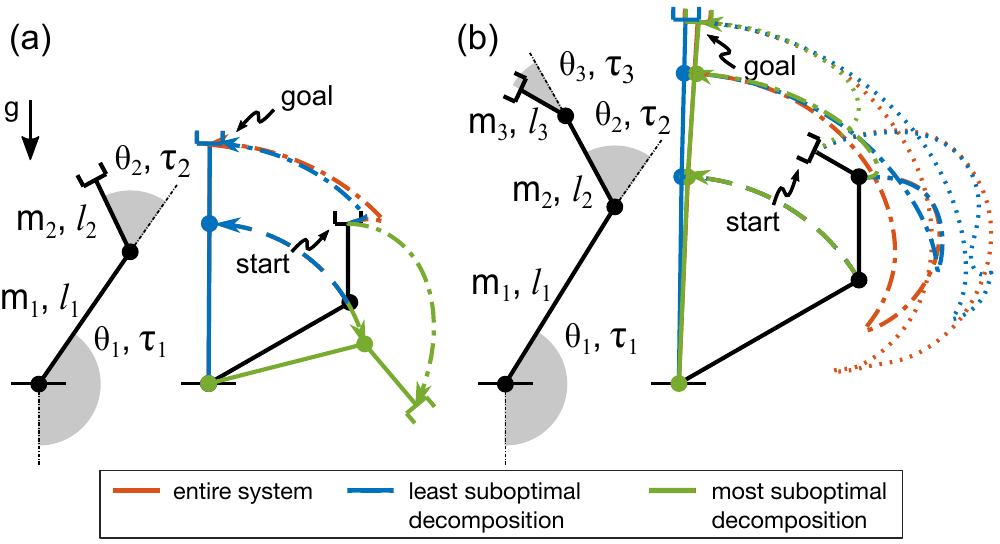}
  \caption{Swing up control for 2, 3 link manipulators. See section~\ref{sec:discussion} and appendix D for details. Behavior of optimal control (red traces) and least and most suboptimal policy decompositions (blue and green) (compare Tabs.~\ref{tab:2linkpendula} and \ref{tab:3linkpendula}).}
  \label{fig:nlinkpendula}
\end{figure}

The 3-link model of a balancing biped shown in Fig.~\ref{fig:biped} has four inputs (leg forces $F_{l/r}$ and hip torques $\tau_{l/r}$) and six states (position defined by leg length $l_r$, leg angle $\alpha_r$, and torso angle $\theta$;  velocities, $\dot{x}$, $\dot{z}$, and $\dot{\theta}$). This system has 110864 pure decompositions (Eq.~\ref{eq:combinations}). But symmetry between the two legs suggests to group leg forces and hip torques into two pseudo-inputs, $\bs{F} = (F_l,F_r)$ and $\bs{\tau} = (\tau_l,\tau_r)$, resulting in just 188 pure decompositions. Additionally, grouping states into pseudo-states for the torso ($\theta$, $\dot{\theta}$) and the center of mass ($l_r$, $\alpha_r$, $\dot{x}$, $\dot{z}$) further reduces this number to 8. Such a grouping lessens the chance for surprise discoveries of controllers but the best among these 8 decompositions yields a policy that has a small value error ($\text{err}^\delta=0.017$, also compare Fig.~\ref{fig:biped}) and is computed about five times as fast as the optimal controller. From Tab.~\ref{tab:bipedresddp}, we see the LQR and DDP estimates correctly identify this best performing policy (appendix C for implementation details). However, decompositions $\#2$ to $\#5$ have large $\text{err}^\delta_{\text{ddp}}$ values despite low true value errors. This is because our strategy to generate initial trajectories for DDP (described in section \ref{sec:DDP}) results in some diverging trajectories for these decompositions. But, the initialization strategy works well for decompositions of all other systems we experimented with.

\begin{table}[t!]
\caption{Actual and predicted closed-loop performance of eight biped model policy decompositions. Input constraints: $0\leq F_{l/r} \leq 3mg$, $|\tau_{l/r}|\leq 0.25 mg/l_0$. 
}
\label{tab:bipedresddp}
\centering
\footnotesize
\begin{tabular}{m{3.1cm}|m{0.3cm}m{0.3cm}m{0.1cm}|m{0.5cm}m{0.2cm}|m{0.65cm}m{0.2cm}}
\textbf{decomposition}& \textbf{time (\%)} & $\textbf{err}^{\boldsymbol{\delta}}$ & \textbf{r} &$\textbf{err}_{\text{lqr}}^{\boldsymbol{\delta}}$ & $\textbf{r}_{\text{lqr}}$ &$\textbf{err}_{\text{ddp}}^{\boldsymbol{\delta}}$ & $\textbf{r}_{\text{ddp}}$ \\
\hline
0: entire system & \scriptsize{100}& \scriptsize{0}&  & \scriptsize{0} & & \scriptsize{0} & \rule{0pt}{3ex}\\
1: $\pi_{\bs{\tau}}\left(\bs{x},\pi_{\textbf{F}}(l_r, \alpha_r, \dot{x}, \dot{z})\right)$ & \scriptsize{19}& \scriptsize{0.017}& \scriptsize{1}& \scriptsize{$7.8\mathrm{e}{-3}$}& \scriptsize{1}& \scriptsize{0.38}& \scriptsize{1} \\
2: $\pi_{\textbf{F}}\left(\bs{x}, \pi_{\bs{\tau}}(\theta,\dot{\theta})\right)$ & \scriptsize{29}& \scriptsize{0.22} & \scriptsize{2} & \scriptsize{$7.9\mathrm{e}{-3}$}& \scriptsize{2}& \scriptsize{440}& \scriptsize{2} \\
3: $\pi_{\textbf{F}}\left(l_r,\alpha_r,\dot{x},\dot{z}\right)$, $\pi_{\bs{\tau}}\left(\theta,\dot{\theta}\right)$ & \scriptsize{0.16}& \scriptsize{0.3}& \scriptsize{3}& \scriptsize{0.016}& \scriptsize{3}& \scriptsize{$1.26\mathrm{e}{3}$} & \scriptsize{3}\\
4: $\pi_{\textbf{F}}\left(\bs{x},\pi_{\bs{\tau}}(\bs{x})\right)$ & \scriptsize{47}& \scriptsize{0.48}& \scriptsize{4}& \scriptsize{0.027}& \scriptsize{4}& \scriptsize{$2\mathrm{e}{3}$}& \scriptsize{4} \\
5: $\pi_{\bs{\tau}}\left(\bs{x},\pi_{\textbf{F}}(\bs{x})\right)$ & \scriptsize{43}& \scriptsize{0.74}& \scriptsize{5}& \scriptsize{0.34}& \scriptsize{6}& \scriptsize{$1.02\mathrm{e}{4}$}& \scriptsize{6} \\
6: $\pi_{\bs{\tau}}\left(\bs{x},\pi_{\textbf{F}}(\theta, \dot{\theta})\right)$ & \scriptsize{21}& \scriptsize{3.17}& \scriptsize{6}& \scriptsize{0.33}& \scriptsize{5}& \scriptsize{$4.19\mathrm{e}{3}$}& \scriptsize{5}\\
7: $\pi_{\textbf{F}}\left(\bs{x},\pi_{\bs{\tau}}(l_r,\alpha_r,\dot{x},\dot{z})\right)$ & \scriptsize{37}& \scriptsize{7.7}& \scriptsize{7}& \scriptsize{4.9}& \scriptsize{7}& \scriptsize{$1.87\mathrm{e}{4}$}& \scriptsize{7} \\
8: $\pi_{\bs{\tau}}\left(l_r,\alpha_r,\dot{x},\dot{z}\right)$, $\pi_{\textbf{F}}\left(\theta,\dot{\theta}\right)$ & \scriptsize{0.15}& \scriptsize{50}& \scriptsize{8}& \scriptsize{$\infty$}& \scriptsize{8} & \scriptsize{$1.85\mathrm{e}{5}$} & \scriptsize{8} 
\end{tabular}
\end{table}

\begin{table}[t!]
\caption{Actual and predicted closed-loop performance of eight 2-link manipulator policy decompositions. Input constraints: $|\tau_{1}| \leq 5$Nm, $|\tau_{2}|\leq 0.5$Nm.}
\label{tab:2linkpendula}
\centering
\footnotesize
\begin{tabular}{m{2.4cm}|m{0.3cm}m{0.6cm}m{0.1cm}|m{0.6cm}m{0.2cm}|m{0.6cm}m{0.2cm}}
\textbf{decomposition}& \textbf{time (\%)} & $\textbf{err}^{\boldsymbol{\delta}}$ & \textbf{r} &$\textbf{err}_{\text{lqr}}^{\boldsymbol{\delta}}$ & $\textbf{r}_{\text{lqr}}$ &$\textbf{err}_{\text{ddp}}^{\boldsymbol{\delta}}$ & $\textbf{r}_{\text{ddp}}$ \\
\hline
0: entire system & \scriptsize{100}& \scriptsize{0}&  & \scriptsize{0} & & \scriptsize{0} & \rule{0pt}{3ex}\\
1: $\pi_{\bs{\tau}_2}\left(\bs{x},\pi_{\bs{\tau}_1}(\Theta_1)\right)$ & \scriptsize{3.5}& \scriptsize{$8\mathrm{e}{-4}$}& \scriptsize{1}& \scriptsize{$2\mathrm{e}{-4}$}& \scriptsize{1}& \scriptsize{$2\mathrm{e}{-4}$}& \scriptsize{1} \\
2: $\pi_{\bs{\tau}_1}\left(\bs{x}, \pi_{\bs{\tau}_2}(\Theta_2)\right)$ & \scriptsize{15}& \scriptsize{$2\mathrm{e}{-3}$} & \scriptsize{2} & \scriptsize{$1\mathrm{e}{-3}$}& \scriptsize{2}& \scriptsize{$1.5\mathrm{e}{-3}$}& \scriptsize{2} \\
3: $\pi_{\bs{\tau}_1}\left(\Theta_1\right)$, $\pi_{\bs{\tau}_2}\left(\Theta_2\right)$ & \scriptsize{0.04}& \scriptsize{$3\mathrm{e}{-3}$}& \scriptsize{3}& \scriptsize{$1.3\mathrm{e}{-3}$}& \scriptsize{3}& \scriptsize{$1.7\mathrm{e}{-3}$}& \scriptsize{3} \\
4: $\pi_{\bs{\tau}_2}\left(\bs{x},\pi_{\bs{\tau}_1}(\Theta_2)\right)$ & \scriptsize{3.8}& \scriptsize{$6.4\mathrm{e}{-3}$}& \scriptsize{4}& \scriptsize{$3\mathrm{e}{-3}$}& \scriptsize{4}& \scriptsize{$0.029$}& \scriptsize{4} \\
5: $\pi_{\bs{\tau}_2}\left(\bs{x}, \pi_{\bs{\tau}_1}\left(\bs{x}\right)\right)$ & \scriptsize{4.8}& \scriptsize{$0.018$}& \scriptsize{5}& \scriptsize{$0.145$}& \scriptsize{5}& 
$4$& \scriptsize{7}\\
6: $\pi_{\bs{\tau}_1}\left(\bs{x},\pi_{\bs{\tau}_2}(\bs{x})\right)$ & \scriptsize{15}& \scriptsize{$0.024$}& \scriptsize{6}& \scriptsize{$1.2$}& \scriptsize{7}& 
\scriptsize{$0.33$}& \scriptsize{5}\\
7: $\pi_{\bs{\tau}_1}\left(\bs{x},\pi_{\bs{\tau}_2}(\Theta_1)\right)$ & \scriptsize{19}& \scriptsize{$0.046$}& \scriptsize{7}& \scriptsize{$0.17$}& \scriptsize{6}& 
\scriptsize{$2.04$}& \scriptsize{6} \\
8: $\pi_{\bs{\tau}_1}\left(\Theta_2\right)$, $\pi_{\bs{\tau}_2}\left(\Theta_1\right)$ & \scriptsize{0.03}& \scriptsize{2}& \scriptsize{8}& \scriptsize{$\infty$}& \scriptsize{8} & 
\scriptsize{$66$}& \scriptsize{8} 
\end{tabular}
\end{table}

\begin{table}[t!]
\caption{Actual and predicted closed-loop performance of five 3-link manipulator policy decompositions. Input constraints: $|\tau_{1}| \leq 16$Nm, $|\tau_{2}|\leq 7.5$Nm, $|\tau_{3}|\leq 1$Nm.}
\label{tab:3linkpendula}
\centering
\footnotesize
\begin{tabular}{m{3.67cm}|m{0.25cm}m{0.45cm}m{0.01cm}|m{0.4cm}m{0.2cm}|m{0.4cm}m{0.2cm}}
\textbf{decomposition}& \textbf{time (\%)} & $\textbf{err}^{\boldsymbol{\delta}}$ & \textbf{r} &$\textbf{err}_{\text{lqr}}^{\boldsymbol{\delta}}$ & $\textbf{r}_{\text{lqr}}$ &$\textbf{err}_{\text{ddp}}^{\boldsymbol{\delta}}$ & $\textbf{r}_{\text{ddp}}$ \\
\hline
0: entire system & \scriptsize{100}& \scriptsize{0}&  & \scriptsize{0} & & \scriptsize{0} & \rule{0pt}{3ex}\\
1: \footnotesize{$\pi_{\bs{\tau}_3}\big(\bs{x},\pi_{[\bs{\tau}_1, \bs{\tau}_2]}(\Theta_1, \Theta_2)\big)$} & \scriptsize{4.6}& \scriptsize{$6.6\mathrm{e}{-3}$}& \scriptsize{1}& \scriptsize{$4\mathrm{e}{-4}$}& \scriptsize{1}& 
\scriptsize{$5.6\mathrm{e}{-3}$}& \scriptsize{1} \\
2: \footnotesize{$\pi_{[\bs{\tau}_1, \bs{\tau}_2]}(\Theta_1, \Theta_2)$, $\pi_{\bs{\tau_3}}(\Theta_3)$}& \scriptsize{0.25}& \scriptsize{$0.051$} & \scriptsize{2} & \scriptsize{$9\mathrm{e}{-4}$}& \scriptsize{2}& 
\scriptsize{$7.6\mathrm{e}{-3}$}& \scriptsize{2} \\
3: \footnotesize{$\pi_{\bs{\tau}_1}\big(\bs{x}, \pi_{\bs{\tau}_2}\big(\Theta_2, \Theta_3, \pi_{\bs{\tau}_3}(\Theta_3)\big)\big)$} & \scriptsize{15}& \scriptsize{$0.094$}& \scriptsize{3}& \scriptsize{$5\mathrm{e}{-3}$}& \scriptsize{3}& 
\scriptsize{$0.053$}& \scriptsize{4} \\
4: \footnotesize{$\pi_{\bs{\tau}_1}(\Theta_1)$, $\pi_{[\bs{\tau}_2, \bs{\tau}_3]}\left(\Theta_2, \Theta_3\right)$} & \scriptsize{0.08}& \scriptsize{0.1}& \scriptsize{4}& \scriptsize{$5.5\mathrm{e}{-3}$}& \scriptsize{4}& 
\scriptsize{$0.082$}& \scriptsize{5} \\
5: \footnotesize{$\pi_{\bs{\tau}_1}\left(\Theta_1\right)$, $\pi_{\bs{\tau}_2}\left(\Theta_2\right)$, $\pi_{\bs{\tau}_3}\left(\Theta_3\right)$}& \scriptsize{$8 \mathrm{e}{-4}$}& \scriptsize{$0.11$}& \scriptsize{5}& \scriptsize{$6\mathrm{e}{-3}$}& \scriptsize{5}& 
\scriptsize{$0.044$}& \scriptsize{3}
\end{tabular}
\end{table}

\new{For the 2 and 3 link manipulators in Fig.~\ref{fig:nlinkpendula} we group angular positions and velocities of each joint $\Theta_i = (\theta_i, \dot{\theta}_i)$, resulting in 8 (enumerated in Tab.~\ref{tab:2linkpendula}) and 180 pure decompositions respectively. To further reduce the possibilities for the 3 link one, we estimate the ratio of policy computation time with and without decomposing. We use grid-based policy iteration for computing policies \cite{Bertsekas:1995} and estimates for computation times can be derived using size of the resulting policy grids, maximum iterations for policy evaluation and update, and number of actions sampled in every iteration for each input. We use these estimates, coupled with the LQR estimate to compute a pareto optimal set of decompositions (Tab.~\ref{tab:3linkpendula}). Even the most suboptimal of these decompositions generates a working policy (Fig.~\ref{fig:nlinkpendula}(b); appendix D)
}

The combinatorics of policy decomposition challenges its practical utility. We only considered pure decompositions in this work, and the problem becomes harder when we consider decompositions that have a combination of decoupling and cascading. Screening decompositions with the LQR estimate and then refining the performance predictions using the DDP estimate may work for moderately complex systems. Using domain knowledge further alleviates this problem but the real test is in identifying promising decompositions for complex and truly unknown systems. Using search methods like GA to prune the possibilities, while accounting for estimates of suboptimality and compute time, is a reasonable future step to achieving this goal.

\section*{APPENDIX} \label{sec:appendix}

\subsection{Pure Decompositions Count} \label{apx:count}
Any decomposition splits the $m$ inputs of a system into $r$ groups, where $r$ can range from 2 to $m$. There are $\Delta(m, r)/r!$ ways to distribute the $m$ inputs into $r$ non-empty groups \cite{Stanley:1986}, with $\Delta(m,r)$ defined in Eq. (\ref{eq:CombDelta}). Thus, the total number of decompositions becomes $N(n,m) =\sum_{r=2}^m \Delta(m,r)/r! \:\:N_{\bs{x}}(n,r)$, where $N_{\bs{x}}(n,r)$ accounts for the number of ways the $n$ states of the system feature in the $r$ input groups. 
For purely decoupled decompositions, $N_{\bs{x}}^{\text{dec}}(n,r) = \Delta(n,r)$, as a particular non-empty $r$-grouping of the state $\boldsymbol{x}$ defines exactly one purely decoupled decomposition. In case of purely cascaded decompositions, sub-policies for an $r$-grouping of inputs can be computed in $r!$ different orders. For each of these orderings a valid $r$-grouping of the state $\bs{x}$ defines a unique cascaded decomposition. A state grouping is valid if the subset of state assigned to the first input group in the cascade is non-empty. There are $(r^n - (r-1)^n)$ valid $r$-groupings of the state. Thus $N_{\bs{x}}^{\text{cas}}(n, r) = r!(r^n - (r-1)^n)$.
Adding $N_{\bs{x}}^{\text{dec}}(n,r)$ and $N_{\bs{x}}^{\text{cas}}(n,r)$ leads to the total number $N(n,m)$ of pure decompositions reported in Eq. (\ref{eq:combinations}).

\subsection{Swing Up control for Cart-Pole}\label{apx:cartpole}
We represent the value functions $V^*(\bs{x})$ and $V^\delta(\bs{x})$ over the state-space,  $\mathcal{S}_{\text{full}} = \{(x, \dot{x}, \theta, \dot{\theta}) \mid x\in[-1.5,1.5], \dot{x} \in [-3,3], \theta \in [0,2\pi], \dot{\theta} \in [-3,3]\}$, with grids of size $31^4$ and use policy iteration \cite{Bertsekas:1995} to compute them. The action-space is $\mathcal{A}_{\text{full}}=\{(F,\tau)\mid|F|\leq6\text{N}, |\tau|\leq 6\text{Nm}\}$. The cost function (Eqs.~\ref{eq:cost} and \ref{eq:obj}) parameters are $\bs{Q} = \text{diag}([25, 0.02, 25, 0.02])$, $\bs{R} = 10^{-3}\text{diag}([1, 1])$, and $\lambda = 3$. We compute $\text{err}^{\delta}$ over a smaller subset $\mathcal{S} = \{(x, \dot{x}, \theta, \dot{\theta})|x\in[-0.5,0.5], \dot{x}\in[-1,1], \theta\in[2\pi/3,4\pi/3], \dot{\theta}\in[-1,1]\}$ to avoid distortions due to state bounds common in grid-based representations. 
\new{For $\text{err}^\delta_\text{ddp}$ (Eq.~\ref{eq:DDPvalueError}), DDP trajectories starting from the 16 states at corners of set $\mathcal{S}$ are computed with time horizon $T = 5$s and time steps $dt = 1$ms.}

\subsection{Balancing control for 3-Link Biped} \label{apx:biped}
Biped (Fig.~\ref{fig:biped}) has a mass $m = 72\text{kg}$, rotational inertia $I=3\text{kgm}^2$, and hip-to-COM distance $d=0.2\text{m}$. Legs are massless and contact the ground at fixed locations $d_f=0.5$m apart. A leg breaks contact if its length exceeds $l_0=1.15$m. In contact,  legs can exert forces ($0\leq F_{l/r}\leq 3mg$) and hip torques ($|\tau_{l/r}|\leq 0.25mg/l_0$) leading to dynamics $m\ddot{x} = F_{r}\cos{\alpha_r} + \frac{\tau_{r}}{l_r}\sin{\alpha_r} + F_{l}\cos{\alpha_l} + \frac{\tau_{l}}{l_l}\sin{\alpha_l}$, $m\ddot{z} = F_{r}\sin{\alpha_r} - \frac{\tau_{r}}{l_r}\cos{\alpha_r} + F_{l}\sin{\alpha_l} - \frac{\tau_{l}}{l_l}\cos{\alpha_l} - mg$, and $I\ddot{\theta} = \tau_{r}(1 + \frac{d}{l_r}\sin(\alpha_r - \theta)) + F_{r}d\cos(\alpha_r - \theta) + \tau_{l}(1 + \frac{d}{l_l}\sin(\alpha_l - \theta)) + F_{l}d\cos(\alpha_l - \theta)$, where $l_l = \sqrt{l_r^2 + d_f^2 + 2l_rd_f\cos{\alpha_r}}$ and $\alpha_l = \arcsin \frac{l_r\sin{\alpha_r}}{l_l}$.

The control objective is to balance the standing biped midway between the footholds. Value functions over state-space $\mathcal{S}_{\text{full}} = \{(l_r, \alpha_r, \dot{x}, \dot{z}, \theta, \dot{\theta}) \mid l_r\in[0.85,1.25], (\alpha_r-\pi/2)\in[0, 0.6], \dot{x}\in[-0.3, 0.5], \dot{z}\in[-0.5,1],\theta \in [-\pi/8,\pi/8], \dot{\theta}\in[-2,2]\}$ are represented with a 6D grid of size $13^2\times 14\times 19 \times 14\times 21$ and computed using policy iteration. The cost function parameters $\bs{Q} = \text{diag}([350, 700, 1.5, 1.5, 500, 5])$, $\bs{R} = 10^{-6}\text{diag}([1,1,10,10])$, and $\lambda = 1$. We compute $\text{err}^\delta$, $\text{err}_{\text{lqr}}^\delta$ and $\text{err}_{\text{ddp}}^\delta$ over a smaller set $\mathcal{S} = \{(l_r, \alpha_r, \dot{x}, \dot{z}, \theta, \dot{\theta}) |l_r\in[0.95,1],(\alpha_r-\pi/2)\in[0.3, 0.4],\dot{x}\in[-0.1,0.1], \dot{z}\in[-0.3,0.3],\theta\in [-0.2,0.2], \dot{\theta}\in[-0.2,0.2]\}$. For $\text{err}^\delta_\text{ddp}$, we compute 64 trajectories starting from the corners of set $\mathcal{S}$, over a horizon of $T=4$s with $dt=1$ms.

\subsection{\new{Swing Up control for Planar Manipulators}} \label{apx:manip}$\text{err}^\delta_{\text{ddp}}$ over set $\mathcal{S}$ is computed similar to the other systems. Time horizon of $T=4$s and time step of $dt=1$ms is used.
\subsubsection*{\textbf{2 DOF}} System dynamics are characterized by $[m_1,m_2] = [1.25,0.25]$kg and $[l_1,l_2] = [0.25,0.125]$m. 
State-space for policy iteration is $\mathcal{S}_{\text{full}} = \{(\theta_1, \theta_2, \dot{\theta}_1, \dot{\theta}_2) \mid \theta_1\in[0,2\pi], \theta_2\in[-\pi,\pi], \dot{\theta}_1,\dot{\theta}_2\in[-3,3]\}$ and value functions are represented using 4D grids of size $31^4$. The action-space is $\mathcal{A}_{\text{full}}=\{(\tau_1,\tau_2)\mid |\tau_1|\leq5\text{Nm}, |\tau_2|\leq0.5\text{Nm}\}$. Cost parameters are $\bs{Q}=\text{diag}([1.6, 1.6, 0.12, 0.12])$, $\bs{R} = \text{diag}([0.003, 0.3])$ and $\lambda=3$. $\text{err}^\delta$, $\text{err}_{\text{lqr}}^\delta$ and $\text{err}_{\text{ddp}}^\delta$ are computed over $\mathcal{S} = \{(\theta_1,\theta_2,\dot{\theta}_1, \dot{\theta}_2)\mid \theta_1\in[2\pi/3,4\pi/3], \theta_2\in[-\pi/3,\pi/3], \dot{\theta}_1, \dot{\theta}_2\in[-0.5,0.5]\}$
\subsubsection*{\textbf{3 DOF}} Masses of the links are $[m_1,m_2,m_3] = [2.75,0.55,0.11]$kg and their lengths are $[l_1,l_2,l_3] = [0.5,0.25,0.125]$m. 
State-space for policy iteration is $\mathcal{S}_{\text{full}} = \{(\theta_1, \theta_2, \theta_3, \dot{\theta}_1, \dot{\theta}_2, \dot{\theta}_3) \mid \theta_1\in[0,2\pi], \theta_2,\theta_3\in[-\pi,\pi], \dot{\theta}_1,\dot{\theta}_2,\dot{\theta}_3\in[-3,3]\}$ and value functions are represented using 6D grids of size $17^3\times13^3$. Action-space is $\mathcal{A}_{\text{full}}=\{(\tau_1,\tau_2,\tau_3)\mid |\tau_1|\leq16\text{Nm}, |\tau_2|\leq7.5\text{Nm}, |\tau_3|\leq1\text{Nm}\}$. The cost parameters are $\bs{R} = \text{diag}([0.004, 0.04, 0.4])$, $\bs{Q}=\text{diag}([1.6, 1.6, 1.6, 0.12, 0.12, 0.12])$ and $\lambda=3$. The set $\mathcal{S} = \{(\theta_1,\theta_2,\theta_3,\dot{\theta}_1, \dot{\theta}_2,\dot{\theta}_3)\mid \theta_1\in[2\pi/3,4\pi/3], \theta_2,\theta_3\in[-\pi/3,\pi/3], \dot{\theta}_1, \dot{\theta}_2, \dot{\theta}_3\in[-0.5,0.5]\}$

\bibliographystyle{IEEEtran}
\bibliography{references}

\end{document}